# INEXPENSIVE SURFACE ELECTROMYOGRAPHY SLEEVE WITH CONSISTENT ELECTRODE PLACEMENT ENABLES DEXTEROUS AND STABLE PROSTHETIC CONTROL THROUGH DEEP LEARNING

Jacob A. George, Anna Neibling, Michael D. Paskett, Gregory A. Clark

*Department of Biomedical Engineering, University of Utah, Salt Lake City, Utah, USA*

**ABSTRACT**

The dexterity of conventional myoelectric prostheses is limited in part by the small datasets used to train the control algorithms. Variations in surface electrode positioning make it difficult to collect consistent data and to estimate motor intent reliably over time. To address these challenges, we developed an inexpensive, easy-to-don sleeve that can record robust and repeatable surface electromyography from 32 embedded monopolar electrodes. Embedded grommets are used to consistently align the sleeve with natural skin markings (e.g., moles, freckles, scars). The sleeve can be manufactured in a few hours for less than $60. Data from seven intact participants show the sleeve provides a signal-to-noise ratio of 14, a don-time under 11 seconds, and sub-centimeter precision for electrode placement. Furthermore, in a case study with one intact participant, we use the sleeve to demonstrate that neural networks can provide simultaneous and proportional control of six degrees of freedom, even 263 days after initial algorithm training. We also highlight that consistent recordings, accumulated over time to establish a large dataset, significantly improve dexterity. These results suggest that deep learning with a 74-layer neural network can substantially improve the dexterity and stability of myoelectric prosthetic control, and that deep-learning techniques can be readily instantiated and further validated through inexpensive sleeves/sockets with consistent recording locations.

## INTRODUCTION

Neural networks have been used to classify hand gestures from surface electromyography (sEMG) with high accuracy [1]. However, these improvements have not been realized for kinematic regression, where the network is used to control multiple degrees of freedom (DOFs) simultaneously and proportionally. Although neural networks are effective at suppressing unintended movement (i.e., reducing cross-talk), their proportional control is noisy, which ultimately can make neural networks inferior to Kalman filters in functional tasks [2]. One explanation for this poor performance is that the amount of data used to train neural networks in past work was roughly two orders of magnitude less than what is traditionally used for deep learning in other domains, and performance is critically dependent on large training datasets [3].

Gathering large datasets of sEMG synchronized to motor intent is particularly challenging because patient time is limited and the placement of recording electrodes changes day to day [4]. Here, we demonstrate a simple approach to gather large datasets of sEMG and thus enable deep learning for myoelectric prostheses. We first introduce an inexpensive sEMG sleeve that can be repeatedly donned with consistent electrode placement, and then we demonstrate how accumulating spatially consistent sEMG over time yields the large datasets necessary for deep learning. The results of this case study suggest that deep learning can improve the dexterity and robustness of myoelectric prostheses.

## METHODS

### Sleeve Design and Fabrication

The sEMG sleeve was constructed from neoprene fabric sewn into a hollow cylindrical shape after electrodes and wires were inserted (Fig. 1A). The neoprene can be stretched during donning and doffing, but also provides enough structural integrity to maintain consistent placement on the forearm. Brass-coated marine snaps served as inexpensive dry electrodes; 32 were embedded across the full circumference and length of the sleeve to record from extrinsic flexors and extensors. Two additional electrodes were embedded at the proximal end of the sleeve to be placed along the ulna bone and serve as an electrical reference and ground. Each electrode was soldered to a segment of flexible wire with high-strength heat shrink to reduce wire breakage. Electrodes were embedded into the neoprene using a crimping tool, and loops of wire were formed to near each electrode to alleviate strain when the fabric is stretched. Wires were stitched onto the neoprene and soldered to a 38-pin SAMTEC connector. Grommets were inserted into the neoprene, unique to one intact individual, such that the grommets aligned with natural skin markings (e.g., freckles, moles, scars) (Fig. 1B). A loose cover made of Lycra® was used to electrically isolate wire and house front-end devices for amplification and filtering (Fig. 1C). The sleeve can be manufactured in a few hours and costs less than $60 (Table 1).

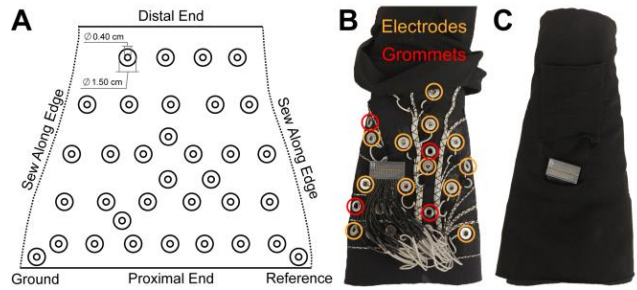

Figure 1: Inexpensive surface electromyography sleeve with consistent electrode placement. A) Thirty-four electrodes (brass-coated marine snaps) are inserted into neoprene, and then the fabric is then sewn into a sleeve. B) Grommets (red) are inserted alongside electrodes (orange) so they align with natural skin markings (e.g., freckles, moles, scars). C) A loose Lycra® cover electrically isolates wire.

### Signal Acquisition

Thirty-two monopolar sEMG electrodes were sampled at 1 kHz using Micro2+Stim Front-Ends and a Grapevine Neural Interface Processor (Ripple Neuro LLC). The 300-ms smoothed Mean Absolute Value (MAV) on the 32 single-ended electrodes (or 528 possible differential pairs) was calculated at 30 Hz [5]. Signal-to-noise ratio (SNR) was defined as the mean 300-ms smoothed MAV during movements (see Training Datasets below) divided by the mean 300-ms smoothed MAV during rest.

Table 1: Cost of Materials

| Component | Description / Use | Cost |
|---|---|---|
| Coated brass marine snap fasteners (34) | Dry recording electrode to record surface electromyography | $8.83 |
| Round stainless-steel washer (34) | Electrode backing to hold electrodes into sleeve | $3.07 |
| Neoprene fabric (1.5 sqft) | Stretch material for easy don/doff | $2.98 |
| Lycra fabric (1.5 sqft) | Cover material to reduce electrical noise with movement/contact | $1.60 |
| Copper wire (26 ft) | Electrode connections | $17.16 |
| Heat shrink (1.5 ft) | Reinforcement for solder joints | $1.62 |
| Thread (3 ft) | Assembly of fabric and wiring | $0.01 |
| SAMTEC connector (1) | Connection to front-end amplification and filtering | $23.20 |
| *Total:* | | **$58.47** |

### Sleeve Performance

Seven intact participants were recruited to validate the sleeve performance. Each participant donned the sleeve five times, attempting to align the grommets with colored markings on their forearms. The times to don and doff the sleeve, as well as the average distance between donned positions, were recorded. Participants also completed one training dataset to determine the sEMG SNR.

### Training Datasets

sEMG and intended movement were recorded simultaneously while participants mimicked the following six preprogrammed movements of a virtual prosthetic hand (MSMS [6]): flexion/extension and abduction/adduction of D1; flexion/extension of (D2); simultaneous flexion/extension D3, D4 and D5; and flexion/extension and pronation/supination of the wrist [5], [7]. Seven intact participants completed one training dataset to determine the sEMG SNR. For one intact participant, a total of 20 datasets were collected over time, each requiring the sleeve to be donned and doffed. For this participant, each dataset introduced slight variations in the movement speed, movement hold-time [2], and forearm posture.

### Neural Network Control Algorithms

Two neural networks were used in this study. The first was a shallow, 10-layer, neural network with similar input and architecture as [2], but with added 50% dropout layers after each rectified linear unit. The second network was a deep, 74-layer, residual neural network with similar architecture as [3]. Input at each timepoint consisted of the MAV from 32 single-ended electrode recordings over the last 32 time-samples (~1.07 seconds) (Fig. 3A). The bulk of the network consists of nine residually connected convolutional units, each of which consisted of two repetitions of a 3x3 convolutional layer followed by batch normalization, followed by a rectified linear unit (Fig. 3B). The output of both neural networks was the kinematic predictions for the six-DOF virtual prosthetic hand. An optional unmodified Kalman filter [5] was placed on the end of deep neural network to smooth the kinematic predictions.

Because tests were performed entirely online, the networks were trained with 97% of the training data, and the remaining 3% was used for validation to avoid overfitting. Training automatically terminated once the root-mean-squared-error (RMSE) on the validation data increased to avoid overfitting. The networks were trained using a

Stochastic Gradient Descent with Momentum solver with an initial learning rate of 0.001 [2].

### Online Performance Metrics

The participant completed a real-time virtual hand matching task in which they actively controlled the virtual prosthetic hand and attempted to move only select DOF(s) to a target location [5]. Performance was evaluated as the mean longest continuous-hold duration (i.e., hold duration) within the 10%-error window around the target location out of a theoretical maximum of seven seconds (i.e., seven seconds max if no reaction time) [5]. Performance of the shallow network, trained only on the first dataset, was evaluated ten times over the span of 263 days. Performance of the shallow network trained on the first ten datasets was also directly compared against the performance of the shallow network trained on only a single dataset collected immediately before the task. Prior results demonstrated that Kalman filters produce smoother movements than neural networks, which is critical for functional tasks [2]. To this end, we evaluated the performance of a deep neural network trained on 20 datasets with and without a Kalman filter to smooth the network output. Direct comparisons were performed using a counterbalanced pseudorandom cross-over design.

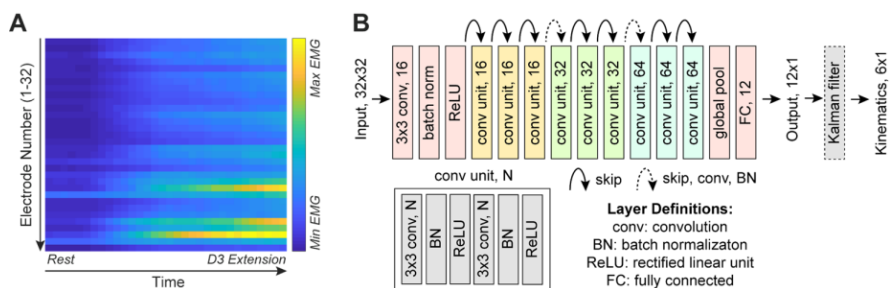

Figure 2: Deep residual neural network for myoelectric prosthetic control. **A)** Example 32-by-32 input "image" consists the 300-ms smoothed MAV on the 32 single-ended electrodes over the last 32 time-samples (i.e., the last ~1.07 seconds). The example image shows EMG activity increasing across all channels as the participant transitions from rest to D3 extension, although EMG activity is highest on channels 23, 28, and 30. **B)** Multiple layers of convolution across electrodes and across time allows for complex non-linear activation patterns. An optional Kalman filter was used to smooth the kinematic predictions from the deep neural network.

## RESULTS

### Inexpensive sleeve enables consistent placement and prosthetic control

Seven intact participants were able to self-don the sEMG sleeve within $7.32 \pm 0.26$ mm of precision in $10.30 \pm 3.35$ s (Fig. 3). The mean SNR for participants was $14.03 \pm 4.43$ (Fig. 3). A shallow neural network, trained on only a single dataset one day before the start of testing, provided relatively stable performance over time in one individual tested longitudinally (Fig. 4A). Performance fluctuated over time ($p < 0.05$, one-way ANOVA), but the performance on day one was not significantly different from that on any subsequent day, including 263 days after initial training ($p$'s $> 0.05$, multiple pair-wise comparisons with correction).

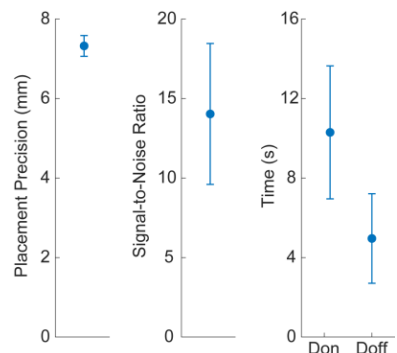

Figure 3: The inexpensive sleeve can be donned rapidly with sub-centimeter precision and adequate signal-to-noise. Data show mean $\pm$ S.E.M. for seven intact participants.

### Deep learning substantially improves prosthetic control

We hypothesized that additional training data would improve neural network performance. To this end, we compared the participant's performance with a shallow neural network trained on a single dataset collected immediately before testing against a shallow neural network trained on 10 datasets collected multiple weeks prior. Performance (hold-time duration) doubled when trained on 10 prior datasets ($p < 0.05$, paired t-test; Fig. 4B).

Deeper neural networks have a greater capacity to learn the intricacies of large complex datasets. Building on this idea, we trained a deep neural network on 20 prior datasets (Fig. 2). Performance significantly improved relative to the shallow networks reported here ($p$'s $< 0.05$, paired t-tests). Deep neural network performance further improved when a Kalman filter was added to the end to smooth kinematic predictions ($p < 0.05$, paired t-test; Fig. 4C). The final architecture, a deep residual neural network with a Kalman filter (DNN+KF), resulted in up to a 152% improvement relative to a modified Kalman filter (4.42 s reported here vs 1.75 s reported previously with the same task and the same participant [2]). Informally, the participant used the DNN+KF to control a physical prosthesis (LUKE Arm) and was able to grasp objects while simultaneously rotating and flexing the wrist – a task that is particularly challenging when simultaneously controlling the position of six different DOFs.

## CONCLUSION

This work first highlights an inexpensive sEMG sleeve with consistent electrode placement. Then, we show how these consistent recordings can enable deep learning, and drastically improve dexterity and stability of myoelectric prosthetic control. Although this latter finding is based on the results from a single participant, the findings are consistent with a recent study that also used spatially consistent sEMG accumulated over time to improve neural networks [8]. Future work should expand this approach to a larger cohort of amputee participants, and validate dexterity and stability in activities of daily living. Additional training paradigms or design considerations may be necessary to account for sEMG variations due to sweat, swelling, and/or excessive fatigue.

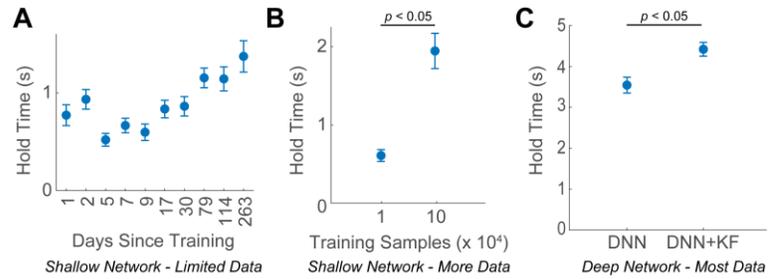

Figure 4: Online performance of prosthetic control. Neural networks were trained to predict motor intent based on surface electromyographic recordings from a custom-made sleeve that maintained consistent electrode placement. **A)** Functional performance, evaluated as the ability to hold complex grasps for up to seven seconds [2, 5], was relatively stable over time. Performance on day one was not significantly different from that on any subsequent day, including 263 days after initial training ($p$'s > 0.05). **B)** A neural network trained on 10 prior datasets accumulated across time doubled the performance of a neural network trained on single dataset from the current day. **C)** A deep neural network (DNN) trained on 20 prior datasets further improved performance relative to the shallow networks in B ($p$'s < 0.05). Adding a Kalman filter to smooth the output of the DNN (DNN+KF) significantly improved performance yet again. DNN+KF performance was 152% greater than what was previously reported for a modified Kalman filter [5] with the same participant [2]. Data show means ± S.E.M. from one participant. p-values shown for paired comparisons.

## ACKNOWLEDGEMENTS

This work was sponsored by NSF Awards GRFP-1747505, ECCS-1533649, and DARPA HAPTIX Contract No. N66001-15-C-4017. Additional support provided by University of Utah Office of Undergraduate Research.